\documentclass[10pt,twocolumn,letterpaper]{article}
\usepackage{booktabs} 
\usepackage{caption} 
\usepackage{tabularx} 
\usepackage{amsmath}
\usepackage{graphicx}
\usepackage{float}
\usepackage{ragged2e}

\usepackage[pagenumbers]{cvpr} 

\usepackage[dvipsnames]{xcolor}

\definecolor{cvprblue}{rgb}{0.21,0.49,0.74}
\usepackage[pagebackref,breaklinks,colorlinks,citecolor=cvprblue]{hyperref}

\def\paperID{7192} 
\def\confName{CVPR}
\def\confYear{2024}

\title{Sculpt3D: Multi-View Consistent Text-to-3D Generation with Sparse 3D Prior}

\author{Cheng Chen$^{1,2}$, Xiaofeng Yang$^{1}$, Fan Yang$^{1}$, Chengzeng Feng$^{1}$, Zhoujie Fu$^{1}$,\\ Chuan-Sheng Foo$^{2,3}$, Guosheng Lin$^{1}$, Fayao Liu$^{2}$\\
$^{1}$Nanyang Technological University \\  $^{2}$Institute for Infocomm Research A*STAR, Singapore \\ $^{3}$Centre for Frontier AI Research, A*STAR, Singapore\\
 \tt\small {cheng021@fudan.edu.cn, {gslin}@ntu.edu.sg {fayaoliu}@gmail.com}}

\begin{document}
\maketitle
\begin{abstract} 
Recent works on text-to-3d generation show that using only 2D diffusion supervision for 3D generation tends to produce results with inconsistent appearances (e.g., faces on the back view) and inaccurate shapes (e.g., animals with extra legs). Existing methods mainly address this issue by retraining diffusion models with images rendered from 3D data to ensure multi-view consistency while struggling to balance 2D generation quality with 3D consistency. In this paper, we present a new framework Sculpt3D that equips the current pipeline with explicit injection of 3D priors from retrieved reference objects without re-training the 2D diffusion model. Specifically, we demonstrate that high-quality and diverse 3D geometry can be guaranteed by keypoints supervision through a sparse ray sampling approach. Moreover, to ensure accurate appearances of different views, we further modulate the output of the 2D diffusion model to the correct patterns of the template views without altering the generated object's style. These two decoupled designs effectively harness 3D information from reference objects to generate 3D objects while preserving the generation quality of the 2D diffusion model. Extensive experiments show our method can largely improve the multi-view consistency while retaining fidelity and diversity. Our project page is available at: \href{https://stellarcheng.github.io/Sculpt3D/}{https://stellarcheng.github.io/Sculpt3D/.}
\end{abstract} 

\section{Introduction} 

\begin{figure}[htbp]
\includegraphics[width=0.48\textwidth]{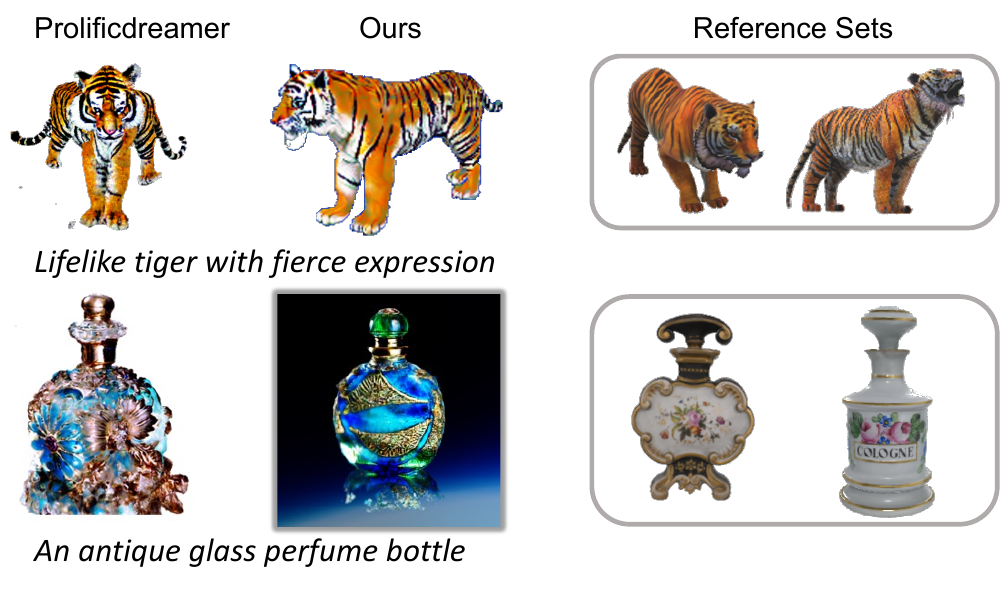}
\caption{Comparison of objects generated by our method and ProlificDreamer. We retain the 2D model's capability to produce high-fidelity objects and adaptively learn 3D information from reference templates retrieved from external datasets.} \label{fig_expl}
\vspace{-0.4cm}  
\end{figure}

\begin{figure*}
\includegraphics[width=\textwidth]{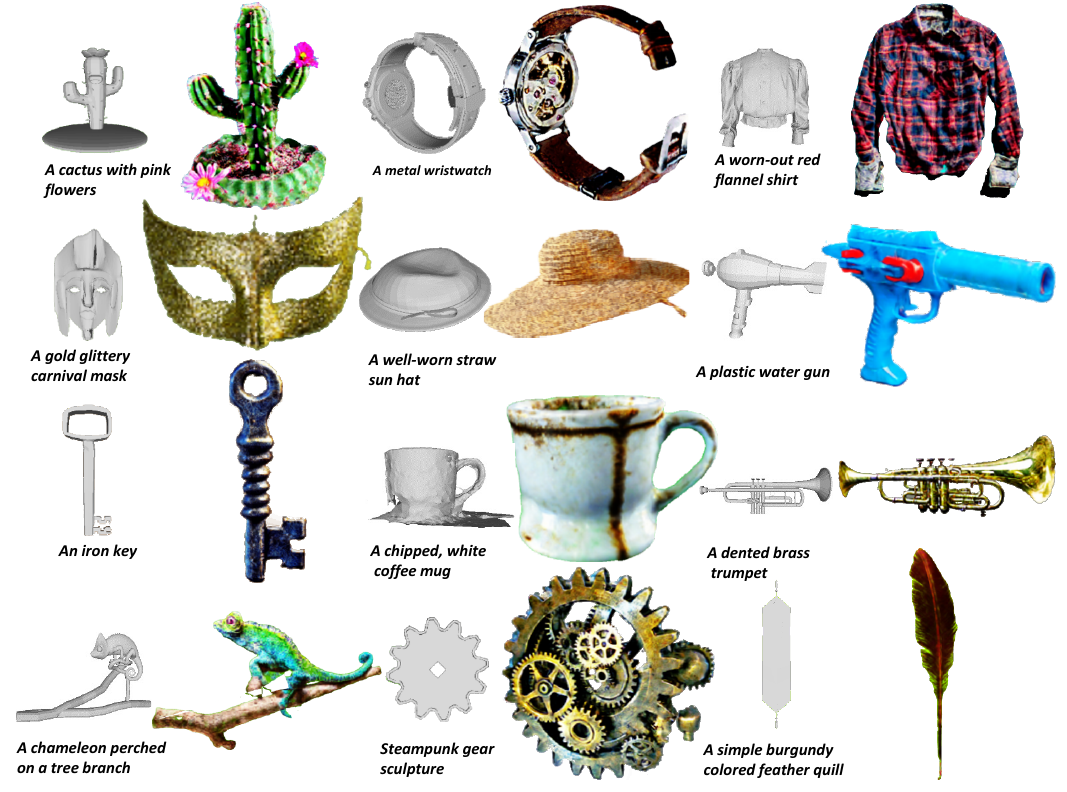}
\caption{Our methods can generate high-fidelity objects with decent shapes using various text prompts. The model adaptively incorporates information from the reference shape displayed on the left, resulting in the creation of objects that range from moderately resembling to substantially diverging from the reference shape. Please find more video results in the supplementary materials.} \label{fig2}
\vspace{-0.4cm} 
\end{figure*}
\begin{figure*}
\includegraphics[width=\textwidth]{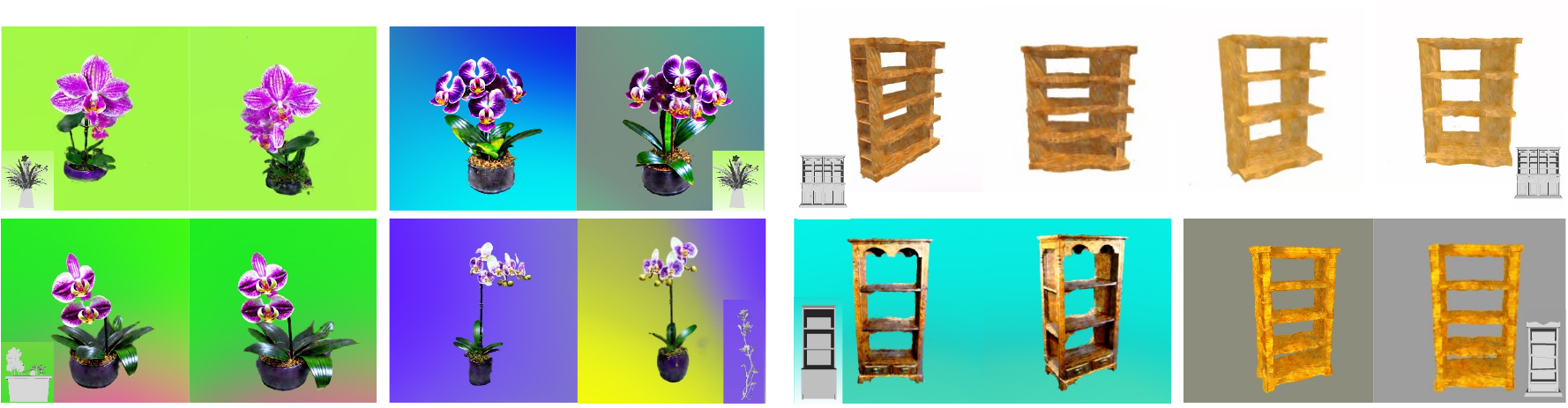}
\caption{As shown in the first row, our method can generate diverse 3D objects given the same reference shape. The second row also shows the diverse results generated by randomly selecting reference objects from the top five retrieved samples. All templates are marked as gray and shown in the corner. } \label{fig3}
\vspace{-0.4cm} 
\end{figure*}

There has been growing research attention towards text-to-3d generation. Compared to image generation, the data available for 3D generation is less in quantity and lower in quality. Thus, many studies \cite{poole2022dreamfusion, lin2023magic3d, wang2023prolificdreamer} have begun to generate 3D objects using 2D text-to-image models \cite{rombach2022high,croitoru2023diffusion} as supervision to leverage their strong priors learned from billions of real images.

These methods mainly contain two steps: the first step is to continuously sample images from different views of a randomly initialized 3D representation (e.g. NeRF \cite{mildenhall2020nerf}, DMTet \cite{shen2021deep}). The second step uses a 2D diffusion model to individually judge whether each image is a high-quality image that conforms to the text description.
Compared to 2D image generation, 3D generation not only requires producing high-quality images for each individual viewpoint but also needs to create plausible shapes and appearances as a whole 3D object. Thus, a high-quality 2D generative model and a mechanism that can accurately provide 3D priors are two keys to achieving decent 3D generation results. Since early works \cite{poole2022dreamfusion, lin2023magic3d, wang2023prolificdreamer}  mainly use the sole 2D diffusion model as supervision, they tend to produce inaccurate shapes (shape ambiguity) and appearances that are inconsistent across viewpoints (appearance ambiguity), as shown in Figure \ref{fig_expl} left, where examples include incomplete bottles, tigers with multiple legs, and tails. 

Recently, some efforts have been made to expand the 3D datasets \cite{deitke2023objaverse}. Following this, there were immediate attempts to retrain 2D diffusion models on these 3D datasets to learn 3D information \cite{liu2023zero, shi2023mvdream, liu2023syncdreamer}. Although these methods have made impressive progress, they require expensive training costs to re-train the large-scale models, and training 2D diffusion models on rendered images often degrades the model's generation quality \cite{liu2023zero, shi2023mvdream} learned on large real image dataset. Similar situations have arisen in the NLP field. As language models grow huger, it becomes increasingly difficult to inject new information by retraining the models, thus researchers start to explicitly introduce external knowledge through retrieval augmentation \cite{shuster2021retrieval, guu2020retrieval}. Motivated by these developments, we design a retrieval mechanism to explicitly supervise the 3D geometry and appearance using retrieved templates without re-training the 2D diffusion model on rendered 3D data.

Explicitly constraining the geometry \cite{song20213d, 10076900} presents an intrinsic challenge: strong constraints on the shape may make the generated results closely resemble the template, while too lax constraints may fail to ensure a reasonable shape \cite{song2023moda}. To adaptively learn the 3D shape information from the template, we exploit the geometric creative capabilities of the 2D diffusion model during volume rendering to enable creative point growth and pruning during the optimization process. Specifically, we design a sparse ray sampling method to selectively discard points, supervising only a minimal number of keypoints that can describe the overall structure, thereby greatly enhancing the 2D diffusion model’s freedom in imaginative shape generation. Moreover, we update the template by pruning and generating new points in areas of low and high NeRF output density, respectively, guided by the diffusion model's confidence.
Since we directly supervise the NeRF without making modifications to the 2D diffusion models, our method can fully preserve the generative quality of the diffusion models while ensuring a decent 3D shape. The generated examples showcased in Figure \ref{fig2} demonstrate that our method is capable of generating photo-realistic objects that adapt to the template shape, with the diffusion model determining the degree of similarity to the template. In cases where users desire results significantly different from the initial template, we further devised a re-retrieval mechanism that corrects the retrieval results through the generated shape to make full use of the external 3D dataset.

The aforementioned design enables our model to generate diverse and accurate 3D objects in most cases. However, we also observed that the diffusion model may still generate appearances that are inconsistent across views despite the shape being accurate. For instance, it may produce the appearance of an animal's face at the back or side view, even when the geometry of the face is not generated there. Thus, we further utilize the template's appearance information to refine the generated objects.

Considering the appearance of generated objects often differs from the template, our challenge here is to correct only the inaccurate aspects of the object's appearance without altering its style and geometry. Fortunately, recent advances in image controlling \cite{zhang2023adding, mou2023t2i, ye2023ip} enable users to easily modify various attributes of an image, such as style, content, and geometry, in a decoupled manner by training lightweight image adapters. Given the fact that the template always provides accurate guidance on view-specific patterns, like which view the eyes and nose should appear. We utilize a unified image adapter to first adapt the template to the generated object's style and then use the adapted image to align the generated erroneous appearances with the correct patterns. As we only modulate the generated patterns for each view without limiting the generated structure, our method only requires four sparse template views to supervise the 3D space partitioned according to four standard orientations. To summarize, our key contributions are:

\begin{itemize}
\item We introduce Sculpt3D which explicitly integrates 3D shape and appearance information for multi-view consistent text-to-3d generation while maintaining the high-quality generation capabilities of the 2D diffusion model.

\item We enable creative point growth and pruning during the 2D diffusion and 3D geometry co-supervision process, which hones 2D diffusion's ability to produce shapes that are both accurate and creative. We further use the appearance pattern information of the template to modulate the output of the diffusion model for resolving appearance ambiguities. 

\item Extensive experiments show that our method is able to significantly improve the multi-view consistency of text-to-3d generation while retaining generalizability.

\end{itemize} 
\begin{figure*}[t]
\includegraphics[width=\textwidth]{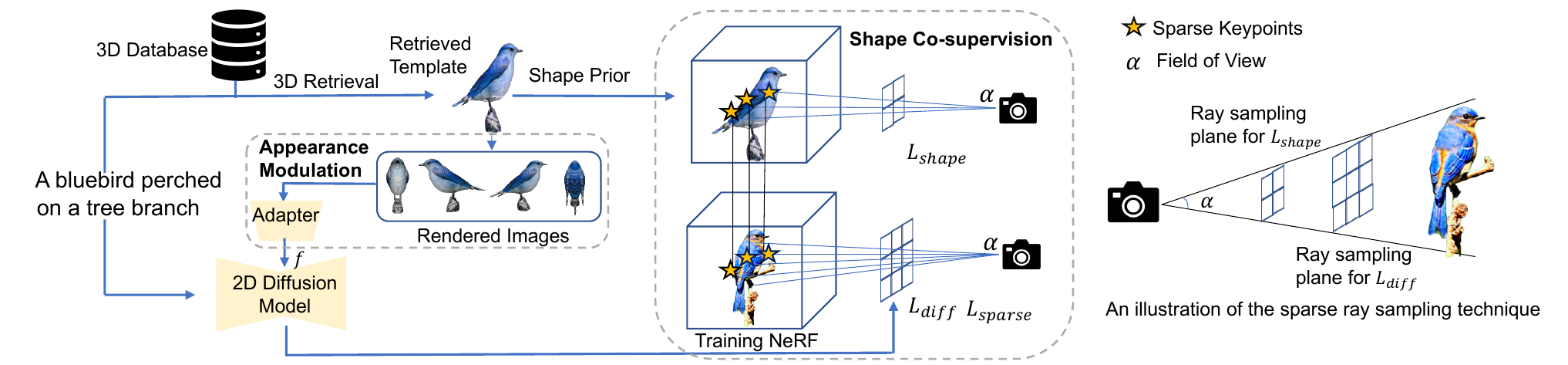}
\caption{Given a text prompt, we retrieve the most semantically matching samples from an external 3D database. With the retrieved object, we sparsely select the keypoints of the reference shape to co-supervise the geometry with 2D diffusion model. The appearance of the reference object is also used to modulate the 2D diffusion to avoid appearance ambiguity. } \label{fig_model}
\vspace{-0.4cm} 
\end{figure*}

\section{Related work} 
\textbf{Large Scale Text-to-Image Diffusion Model.}  With the tremendous progress in large-scale generative models, a surge of methods \cite{rombach2022high,croitoru2023diffusion}  have been proposed to perform various types of text-to-image Generation and Editing. To further enhance the generative capabilities of large models, various methods \cite{ruiz2023dreambooth, hu2021lora} have been proposed to integrate external control signals into these models. ControlNet \cite{zhang2023adding} fine-tunes the Stable Diffusion \cite{rombach2022high} models to enable more conditional inputs like edge maps, segmentation maps, keypoints, etc. Similar to ControlNet, T2I-Adapter \cite{mou2023t2i} and IPAdapter \cite{ye2023ip} introduce lightweight adapters for different conditions, providing additional conditional control and supporting the simultaneous use of multiple conditions for one generation. Using external knowledge to augment models has recently drawn attention in both NLP and visual models \cite{borgeaud2022improving, lewis2020retrieval, chen2022re, blattmann2022retrieval, chen2022utc}. In image synthesis, Re-Imagen \cite{chen2022re} retrieves semantic neighbors to improve the grounding of the diffusion models to real-world knowledge. RDM \cite{blattmann2022retrieval} empowers smaller models with external memory to achieve high-fidelity image generation results. Inspired by these approaches, we utilize template appearances as references to modulate the diffusion process, ensuring the generated images align with the intended viewpoints.

\textbf{Learning 3D from 2D Diffusion Prior.}
Pioneering works like Dreamfusion \cite{poole2022dreamfusion} and SJC \cite{wang2023score} demonstrate the possibility of supervising NeRF to generate 3D objects using only 2D diffusion. Although their advancements are groundbreaking, the results they produced are somewhat blurry. Subsequent researchers approach the challenge from various perspectives \cite{yang2023learn, chen2023gaussianeditor, chen2023it3d, zhang2023styleavatar3d}. Specifically, Magic3D \cite{lin2023magic3d} improves both the speed and quality by introducing DMTet \cite{shen2021deep}. Latent-NeRF \cite{metzer2023latent} seeks to optimize NeRF from an implicit space perspective. Fantasia3D \cite{chen2023fantasia3d} separates geometry and texture modeling to better learn the details of 3D objects. Prolificdreamer \cite{wang2023prolificdreamer} introduces the VSD loss to learn the variational distribution of 3D scenes, significantly improving the generation quality. There's also a line of works focused on image-to-3D generation. For instance, several works \cite{melas2023realfusion, deng2023nerdi} use image-conditioned diffusion as a prior to enhancing the generation of unseen viewpoints. 

With the release of the large 3D dataset \cite{deitke2023objaverse}, recent methods \cite{liu2023syncdreamer, long2023wonder3d, shi2023zero123++, shi2023mvdream} have attempted to fine-tune 2D diffusion with 3D data. Among them, Zero 1-2-3 \cite{liu2023zero} introduces camera parameters as conditions to predict images from arbitrary angles relative to the input image. MVDream \cite{shi2023mvdream} proposes 3D self-attention to further enhance the generation. Syncdreamer \cite{liu2023syncdreamer} synchronizes the multiview diffusion model to produce multiple new viewpoint images simultaneously.

Different from previous works, our Sculpt3D explicitly explores the 3D priors from reference samples to enhance both the generated shape and appearance without retraining the diffusion model.

\section{Approach} 
As shown in Figure \ref{fig_model}, our method uses retrieved templates to provide shape and appearance priors for shape co-supervision and appearance modulation. The details of these components will be given in the following sections.

\subsection{Revisiting 2D Diffusion for 3D Generation}
Dreamfusion \cite{poole2022dreamfusion} introduces a Score Distillation Sampling (SDS) loss to perform text-to-3d generation. The loss is designed for distilling knowledge from 2D diffusion models to train a 3D representation. Specifically, given a NeRF model $g(\theta)$ which can produce image $x$ at arbitrary camera poses, SDS provides the gradient direction to update $\theta$ such that all rendered images are pushed to the high probability density regions conditioned on the text embedding $y$ under the diffusion prior. The SDS computes the gradient as:
\begin{equation}\resizebox{.9\hsize}{!}{$
\nabla_{\theta} \mathcal{L}_{\mathrm{SDS}}(\phi, x=g(\theta)) = \mathbb{E}_{t,\epsilon} \left[ w(t) \left( \epsilon_{\phi}(z_t;y,t) - \epsilon \right) \frac{\partial x}{\partial \theta} \right],$}
\end{equation}

where $w(t)$ is a weighting function, $z_t$ is the noised latent of image $x$ at timestep $t$, and $\epsilon_\phi$ is the denoising network of Stable Diffusion. As aforementioned, while the SDS loss can effectively train the NeRF model, its generated outputs often suffer from oversaturation and are lacking in detail.

To address these issues, Prolificdreamer \cite{wang2023prolificdreamer} introduces the VSD Loss. The VSD Loss incorporates the LoRA \cite{hu2021lora} model to further fit the variational distribution of the 3D scene produced by the training NeRF. It then computes the difference between a pre-trained diffusion model and the LoRA model to guide the NeRF, which is formulated as:
\begin{align}
\nabla_{\theta} \mathcal{L}_{VSD}(\theta) &\triangleq \mathbb{E}_{t,\varepsilon,c} \left[ \omega(t) \cdot \left( \epsilon_{\text{pretrain}}(x_t, t, y) \right. \right. \notag \\
&\hspace{1.5cm} \left. \left. - \epsilon_{\phi}(x_t, t, c, y) \right) \cdot \frac{\partial g(\theta, c)}{\partial \theta} \right]. \label{eq:combined}
\end{align}

In the formula, $\epsilon_{\phi}$ represents the score of a noisy rendered image predicted by the LoRA model, and $c$ is the camera parameter corresponding to the rendered view. We recommend readers refer to Prolificdreamer's \cite{wang2023prolificdreamer} original paper for more details. The VSD loss can effectively improve the fidelity of the generated samples, thus we use it as the 2D diffusion prior by default.

In our experiments, we found that although the VSD loss is able to produce detailed results, its outputs still suffer from inaccurate shapes and appearances. To address these challenges, we next introduce our method of equipping the current pipeline with retrieval capability to explicitly inject 3D priors in the following sections.

\subsection{3D and 2D Co-supervised 3D Generation}
Based on previous observations, we now turn to illustrate how to use 3D prior when doing text-guided 3D generation. In our setup, 3D priors can be obtained either by user input or retrieved from external datasets. Recent advances in representation learning suggest that by scaling up 3D representations, it is accessible to align the CLIP \cite{radford2021learning} space with 3D data, thereby enabling the retrieval of the most semantically matching objects in a 3D database using natural language. In the experiment, we use the recently released OpenShape \cite{liu2023openshape} model which scales up the 3D backbone to align with CLIP as our 3D retrieval module. NeRF is chosen as our 3D representation due to its flexibility in modifying prior shapes.

To inject the 3D prior, we initially used the 3D template shape to directly initialize the volume density of NeRF.
Specifically, inspired by Latent-NeRF \cite{metzer2023latent}, we constrain the density of each point during NeRF training. Here, the density label of each point is calculated from the winding number \cite{barill2018fast} of the normalized template. If the winding numbers show that a point is inside the 3D template shape, we set the density label of that point to 1. Conversely, the point's label is set to 0. After obtaining an accurate initialized shape, we continue training NeRF using 2D diffusion as supervision. However, we find that the 2D diffusion tends to destroy the initial object shape and converge to a distorted shape. Similar results are also observed by \cite{haque2023instruct, shao2023control4d}. They find that continuing to modify a well-trained NeRF using either SDS or VSD loss will destroy the initially well-learned 3D representation. In order to effectively generate the correct geometry, we supervise NeRF using both 2D diffusion and 3D shapes. Which is formulated as:
\begin{equation}\label{eq3}
   L_{co} =  L_{{diff}} + \lambda L_{shape}.
\end{equation}

As aforementioned, too tight supervision on the shape will make the generated result too similar to the template, and sometimes it even produces an incorrect appearance due to discrepancies between the diffusion prior and the template shape prior. To effectively use 3D prior, we next introduce our shape learning method.

\subsubsection{3D Prior Guided Shape Learning}
\label{sec:method_shape_learning}
To allow the diffusion model to adaptively learn the 3D prior, we introduce a sparse ray sampling technique to selectively supervise a small number of keypoints that roughly describe the object's shape. Specifically, every time when randomly sampling a view to train NeRF, 2D diffusion is utilized to supervise all rays to learn the correct RGB and density of each point in the 3D space. At the same time, we maintain the field of view (FOV) unchanged and proportionally reduce the width and height of the ray sampling plane by a factor of N for shape supervision. In this way, as shown in Figure \ref{fig_model}, the sampled rays are much sparser, roughly depicting the 3D object shape and providing direct shape guidance.
Since the shape constraint only provides a correct sparse prior, diffusion can freely unleash its generative capabilities in the unconstrained space. The shape supervision loss is defined as a binary cross-entropy loss:
\begin{equation}
    L_{shape}= - \frac{1}{|\cal R|}\sum_{o \in  {\cal R} } [s_o \log d_o + (1-s_o) \log (1-d_o)],
\end{equation}
where $\cal R$ denotes the set of keypoints, $s_o$ denotes the density of the keypoint $o$, and $d_o$ is the NeRF output density. 

Considering the co-supervision of 3D shapes and 2D diffusion, two types of conflicts may arise: 2D diffusion might tend to either prune certain points existing in the template or generate points that are not present in the template. To leverage the creativity of the diffusion model to drive the model's generation, we default the shape supervision scale $\lambda$ in equation \ref{eq3} to 0.1. With this configuration, as the diffusion loss scale is larger, points prone to pruning by the diffusion model will have their density optimized towards 0. Conversely, points inclined to be generated will be optimized towards 1. To accelerate the removal of these unnecessary points and the growth of new ones, we further impose a sparsity loss to enforce the generated points' density to be either zero or one, and the points with a density of zero will be pruned and no longer supervised. As shown in the results Figure \ref{fig2}, our technique can effectively remove unwanted points and generate new points to create new shapes. The sparsity loss is defined as follows:
\begin{equation}
    L_{sparse} = \frac{1}{|\cal T|} \sum_{ o \in \cal T} [\log(\ d_o ) + \log(1 -  d_o )],
\end{equation}
where $\cal T$ denotes the set of all points. The final loss we used is $L=L_{co}+L_{sparse}$.

The aforementioned design can effectively assist the model in generating new shapes. When the user wishes to significantly increase the downsampling factor $N$ to create objects that differ greatly from the template, we further design a re-retrieval mechanism. Specifically, we extract the initially generated shape representation and use it to retrieve matching shapes in the top 100 objects retrieved by text. This allows for further utilization of 3D datasets to find the reference shape that best matches the structure generated by diffusion.

\subsubsection{3D Appearance Modulated 2D Diffusion Prior}
Couple shape guidance with 2D diffusion prior can effectively help the model to correctly understand the 3D world, thereby producing correct generation results. However, in our experiments, we also find that the model still cannot infer the correct appearance even when the shape is entirely accurate in some hard cases. To explicitly guide the model to generate the correct appearance for each view, we design an optional technique that uses the appearance of the template as a semantic reference to modulate the diffusion process in hard cases.

\begin{figure}[h]
\includegraphics[width=0.48\textwidth]{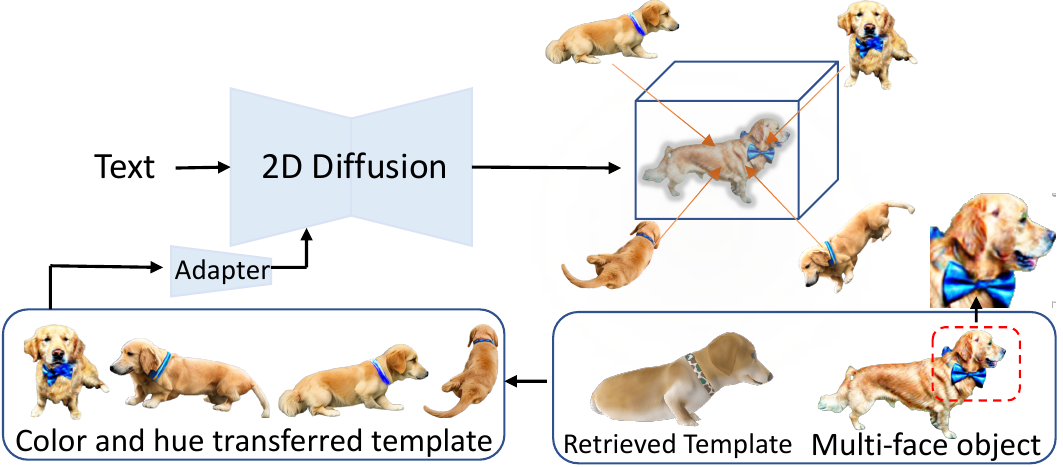}
\caption{Illustration of the appearance modulation. Four canonical views of the templates are transferred to the generated object's style to modulate the 2D diffusion.} \label{fig_al}
\vspace{-0.4cm} 
\end{figure}

Specifically, recent works \cite{mou2023t2i, ye2023ip} demonstrate that various external signals can be applied to control the output of diffusion by training a lightweight adapter. The adapter can utilize images as prompts to generate results with semantic patterns similar to the reference image. In order to correct the appearance of the generated object without affecting its overall style, we first utilize the adapter to convert the template objects to match the hue and color distribution of the generated one. This can be simply achieved by using the color distribution of the generated object as a condition. Since the converted template view contains accurate view-specific patterns, it is used as the image prompt together with the text to align the diffusion generation results with the correct semantics pattern. 
As LoRA is designed to fit the scene distribution of the trained NeRF in VSD loss, the diffusion prior coupled with the image adapter can be formulated as:

\begin{align}
\nabla_{\theta} \mathcal{L}_{diff}(\theta) &\triangleq \mathbb{E}_{t,\varepsilon,c} \left[ \omega(t) \cdot \left( \epsilon_{\text{pretrain}}(x_t, f, t, y) \right. \right. \notag \\
&\hspace{1.5cm} \left. \left. - \epsilon_{\phi}(x_t, t, c, y) \right) \cdot \frac{\partial g(\theta, c)}{\partial \theta} \right], \label{eq_am}
\end{align}
where $f$ denotes the image features extracted by the adapter. As shown in figure \ref{fig_al}, since the semantic pattern of the image is constant within a certain observation range, our method only requires 4 sparse template images corresponding to 4 canonical view spaces.

\begin{figure*}[h]
\includegraphics[width=\textwidth]{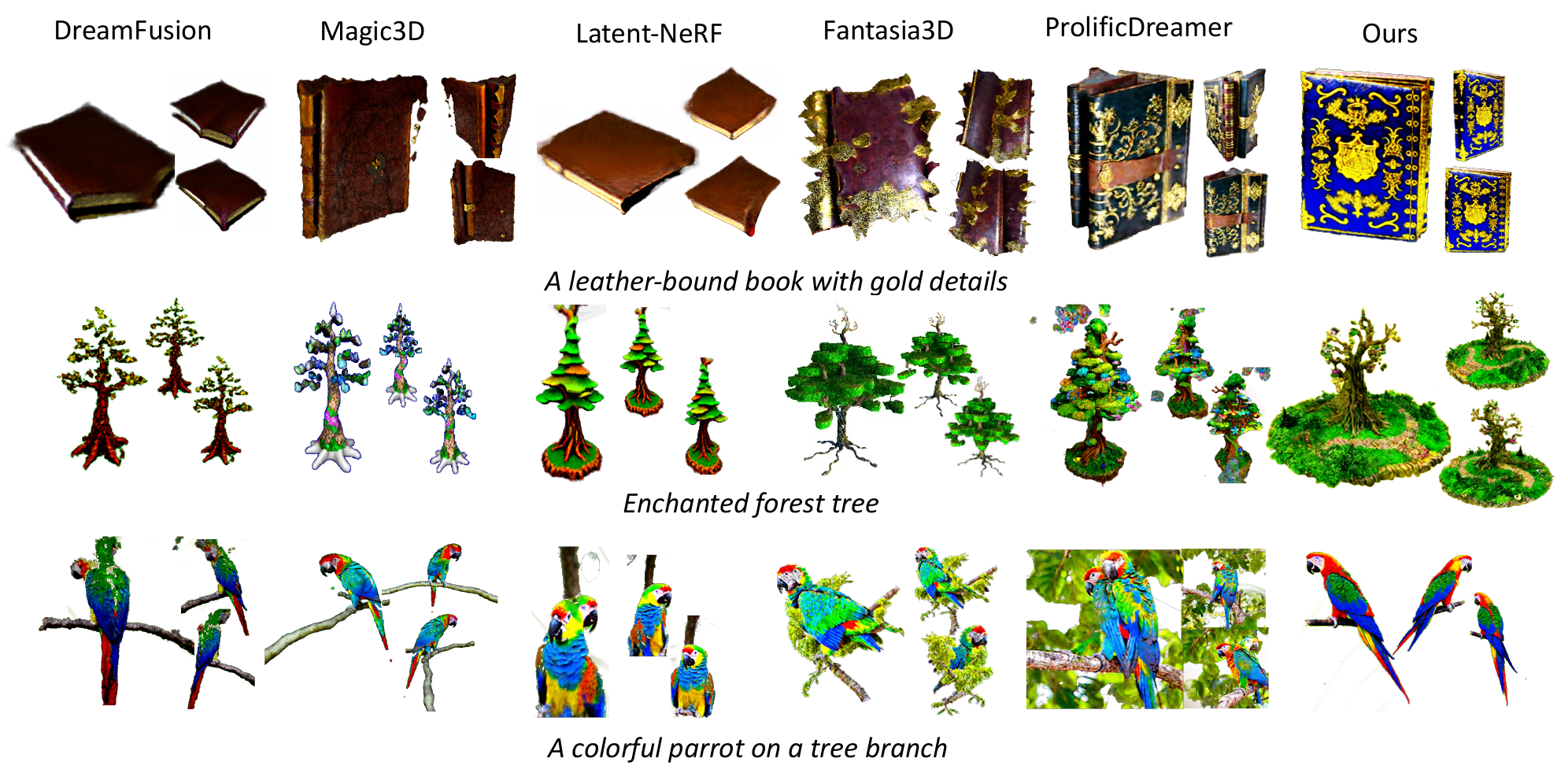}
\caption{Comparison with baselines. Our method can generate objects with decent shapes, which not only have high fidelity and rich details but also maintain 3D consistency.} \label{fig_compare}
\vspace{-15pt}
\end{figure*}

\section{Experiments}
To comprehensively assess the effectiveness of our method, we use the text descriptions provided by T3Bench \cite{he2023t} for testing, which contains 100 text prompts covering various types of single objects. Additionally, we use ChatGPT to generate 40 different prompts for testing, including both common everyday objects and some imaginative objects. 

\begin{figure}[h]
\includegraphics[width=0.48\textwidth]{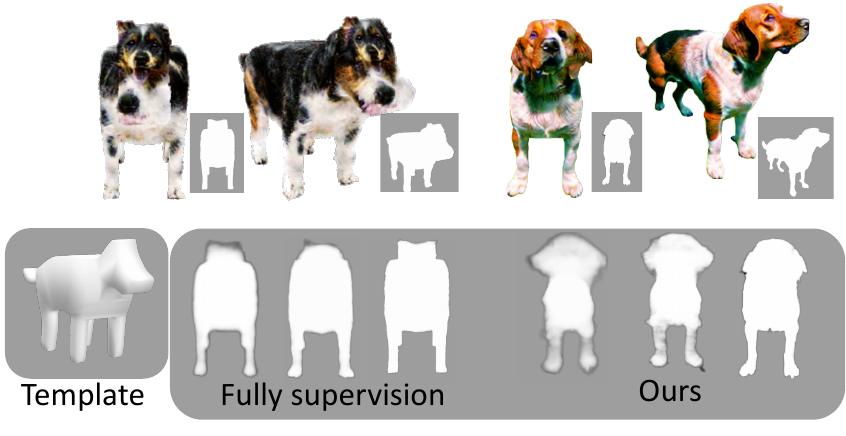}
\caption{Comparisons with fully shape supervision, the density changes during training are depicted in the gray box below. It is observable that full supervision tends to result in the generated shapes closely resembling the template, thereby leading to a loss in diversity. Moreover, the discrepancy between 2D priors and 3D shape priors could potentially result in inaccuracies in the appearance of the shapes.} \label{fig_shape_ab}
\vspace{-15pt}
\end{figure}

\subsection{Implementation Details}
Our method is built on the implementation from Threestudio \cite{threestudio2023}. All experiments are conducted on an NVIDIA A6000 GPU. The model used for 3D retrieval is OpenShape \cite{liu2023openshape}, an open-world retrieval model trained using multiple ensemble datasets. By default, the shape ranked first in the retrieval results is used as the reference shape in the experiments. The scale of constraint on shape $\lambda$ is consistently set to 0.1. The sparse keypoints selection factor N is set to 8, meaning 2D diffusion supervises 3D points on rays sampled from a 512×512 space, while geometry supervision is applied to 3D points on rays sampled from a 64×64 image space. The initial shape is obtained at the 5000 training step when performing shape retrieval. The version of diffusion used in the experiments is Stable Diffusion 1.5. The image adapter used in appearance learning is the publicly available pre-trained T2I-Adapter \cite{mou2023t2i}.

\subsection{Results of Sculpt3D}
We show the generated results of Sculpt3D in Figure \ref{fig2}, including the generated results and the corresponding reference shapes shown on the left. The results demonstrate that our method can generate objects with accurate geometry using various text descriptions while maintaining the ability of 2D diffusion to produce highly realistic appearances. It can be observed that our method can adaptively learn geometry information from the template. Some generated objects resemble the reference shapes, while others show significant differences. Moreover, Figure \ref{fig3} also showcases Sculpt3D's capability to produce diverse results.
We illustrate this through two sets of examples: one where the same template is used to generate multiple times and another where various templates are randomly chosen from the top five retrieved results. The results are shown in the first and second rows respectively. It can be seen that even when using the same template, Sculpt3D can produce remarkably different results due to the sufficient creative freedom allowed in the generation process. Additionally, since the external 3D dataset contains different samples that conform to the same semantic description, randomly selecting from the retrieval results also effectively yields diverse generative results.
\begin{figure}[h]
\includegraphics[width=0.48\textwidth]{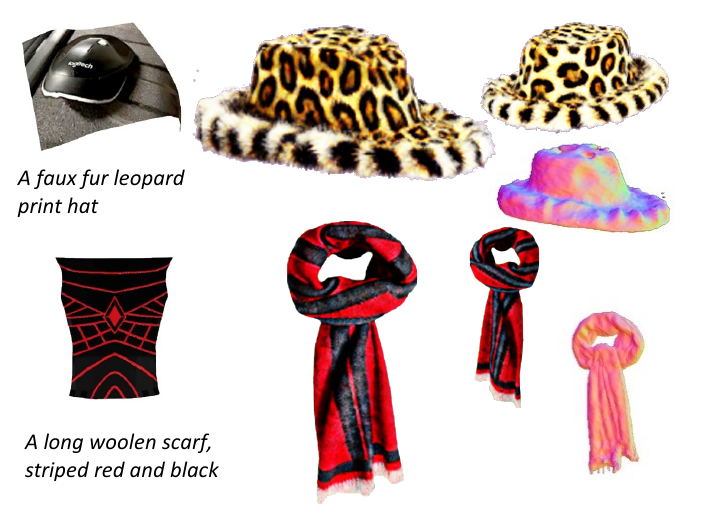}
\caption{We also showcase the generation results when no matching samples are retrieved. Even though the retrieval model failed to find samples that fully match the semantics, our method is still capable of effectively absorbing useful information from the template to produce correct results.} \label{fig_shape_fai}
\vspace{-0.5cm} 
\end{figure}
\subsection{Comparison to Baselines}
We compare our method with five baselines, DreamFusion \cite{poole2022dreamfusion}, Latent-NeRF \cite{metzer2023latent}, Magic3D \cite{lin2023magic3d}, Fantasia3D \cite{chen2023fantasia3d}, and ProlificDreamer \cite{wang2023prolificdreamer}. We use the implementation from Threestudio \cite{threestudio2023} for all these baselines. As shown in Figure \ref{fig_compare}, previous methods struggle to generate shapes with reasonable geometry and high-quality appearances, while our method is capable of simultaneously producing objects with good geometry, higher fidelity, and more details.

\begin{figure}[h]
\includegraphics[width=0.48\textwidth]{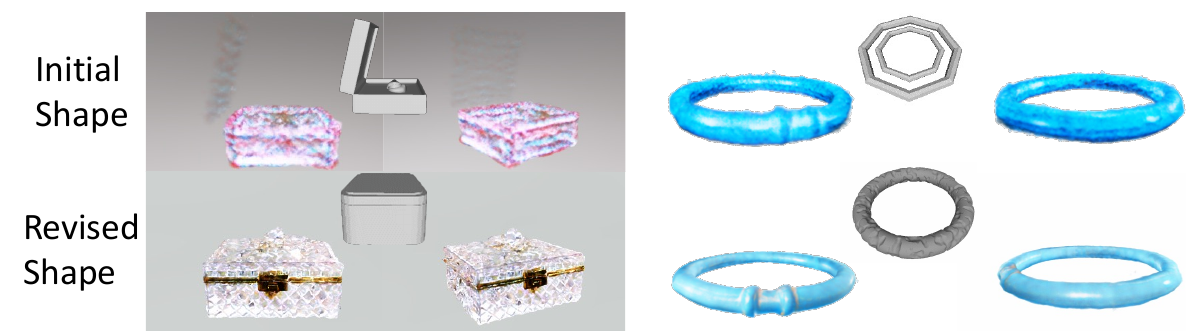}
\caption{Illustration of the effectiveness of re-retrieval using generated shapes. When users desire results significantly different from the initial template, we can utilize the outputs from the rough generation stage, as shown in the first row, to re-retrieve more accurate reference shapes, demonstrated in the second row.} \label{fig_re}
\vspace{-0.2cm} 
\end{figure}

\begin{table}[hb]
\setlength{\tabcolsep}{10.0pt}
\fontsize{9}{10}\selectfont
\centering
\caption{Quantitative comparisons with the baselines.}\label{tab1}
\begin{tabular}{lcccc}
\toprule
Methods&Quality&Alignment&Cons. Rate\\ 
\midrule
Dreamfusion & 24.9 & 24.0 & 34\% \\
Latent-NeRF & 34.2 & 32.0 & 30\%  \\
Magic3D & 38.7 & 35.3 & 38\% \\
Fantasia3D & 29.2 & 23.5 & 26\% \\
ProlificDreamer & 51.1 & 47.8 & 32\% \\
\textbf{Ours} & \textbf{53.6} & \textbf{49.3} & \textbf{76\%}\\
\bottomrule
\end{tabular}
\vspace{-10pt} 
\end{table}

\subsection{Quantitative Evaluation}
To quantitatively evaluate the text-to-3d method, T3Bench \cite{he2023t} designs two metrics based on multi-view CLIP score and GPT-4 evaluation to assess the generated object's quality and alignment using 100 prompts. As our work focuses on generating 3D content with multi-view consistency, we further follow previous work \cite{li2023sweetdreamer} to evaluate 3D consistent rate. Specifically, we randomly select 50 prompts from T3Bench, and manually identify and count 3D inconsistencies (e.g., multiple faces, legs, and other distorted shapes.) of each method. The consistent rate is then determined by dividing the number of 3D consistent objects by the total generated results. As shown in Table \ref{tab1}, our method significantly improves the multi-view consistency rate while surpassing the baseline in both quality and alignment metrics.

\subsection{Ablation Study}
\label{sec:method_ablation}
\begin{figure}[h] \centering
\includegraphics[width=.9\columnwidth]{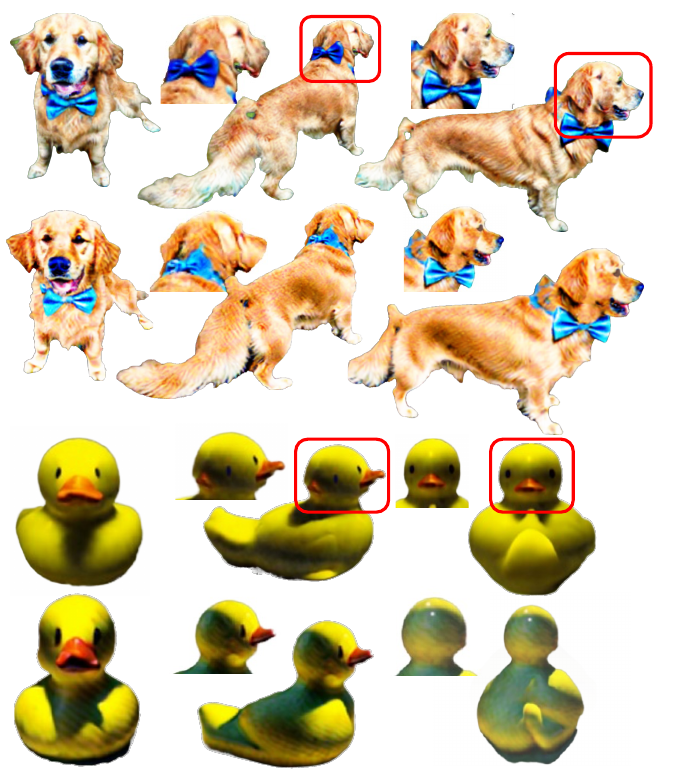}
\caption{Multi-view inconsistency (appearance) issue. Each column shows a single view of the object. \textbf{1st, 3rd rows}: Despite the accuracy of the shape, there is a potential for ambiguity in appearance (highlighted in the red box) that may still arise. \textbf{2nd, 4th rows}: Our appearance modulation method can effectively correct this type of appearance confusion.} \label{fig_ap_refine}
\vspace{-8pt}
\end{figure}

\textbf{Effectiveness of Shape Learning.} 
As shown in Figure \ref{fig_shape_ab}, we compare the progression of NeRF density changes and the final generated results of our method with the fully shape supervision method proposed in Latent-NeRF \cite{rombach2022high}, which is designed to make minor modifications to a given shape surface. The result shows that when applying their constraining coupled with VSD on the high-resolution NeRF, the model converges quickly to the reference shape, excessively limiting the diffusion model's creativity and resulting in an unnatural shape. Furthermore, the gap between the diffusion model's prior and the template's geometry may cause the model to generate incorrect appearances in inappropriate locations (e.g., multiple dog faces). In cases where the reference shapes are not accurate, it becomes increasingly crucial to grant the diffusion model a suitable level of creative freedom. In Figure \ref{fig_shape_fai}, we show the generated results when the mismatching templates are retrieved. It is evident that our model can still effectively generate correct results by adding and pruning points based on the inaccurate template shape. 
We also demonstrate situations where further reducing the supervised points. Figure \ref{fig_re} illustrates the scenario when the sparse selection factor is set to 16. In this scenario, the model may make significant modifications to the initial reference shape, such as removing the reference shape's lid or combining two rings of a bracelet into one. Our re-retrieval method still can effectively utilize the initial generated shape to retrieve the most matching sample from the candidate set, thus accelerating the creation of the final high-quality object.

\textbf{Effectiveness of Appearance Modulation.} Though the correct shape is guaranteed by shape supervision, some challenging cases also show appearance ambiguities. As shown in the first row of Figure \ref{fig_ap_refine}, our baseline model can generate the correct shape of a yellow rubber duck, but behind its head, despite the absence of corresponding shapes for the mouth and eyes, it still generates the appearance of a duck's face at the back view. By using the template with the correct pattern to refine erroneous appearances, our method can effectively adapt the generated objects to the correct appearance without changing their overall style. In the case of the golden retriever, which presents a difficult pose, our baseline model initially generates an extra face on the side by mistake. Through refinement, we are able to effectively correct the appearance mismatch for this challenging pose.

\section{Conclusion \& Limitation}
\textbf{Conclusion:} In this paper, we propose Sculpt3D which explicitly utilizes the 3D shape and appearance information from the retrieved template to aid text-to-3D generation. Sculpt3D is capable of performing creative point growth and pruning within a framework of sparse geometry constraints, thus enabling flexible and accurate shape generation. Moreover, we use the correct pattern information from the template's appearance to fix ambiguities in the generated object's appearances without changing their style. Sculpt3D enhances multi-view consistency in a manner that explicitly supervises the 3D representation, thereby preserving the generative capabilities of 2D diffusion. Experiments on text-to-3d benchmarks show the effectiveness of the proposed model, and more extensive ablation studies further confirm the generalizability of our method.

\textbf{Limitation:} While our method has shown promising performance, we also note some limitations. Since we explicitly supervise the geometry, it is difficult for our method to correctly generate when the initial retrieved shape exceeds the prior of the 3D dataset. Early in our development, we experimented with generating an initial 3D shape without constraints and then using that shape to retrieve matching reference objects. Unfortunately, this approach often fails to produce reasonable reference objects due to the limitations of existing generation methods.

\justify
\textbf{Acknowledgements.} This research work is supported by the Agency for Science, Technology and Research (A*STAR) under its MTC Programmatic Funds (Grant No. M23L7b0021).
\endjustify
{
    \small
    \bibliographystyle{ieeenat_fullname}
    \bibliography{main}
}

\clearpage

	\clearpage
	
	\renewcommand\thesection{\Alph{section}}
	\renewcommand\thesubsection{\thesection.\arabic{subsection}}
	\renewcommand\thefigure{\Alph{section}.\arabic{figure}}
	\renewcommand\thetable{\Alph{section}.\arabic{table}} 
	
	\setcounter{section}{0}
	\setcounter{figure}{0}
	\setcounter{table}{0}
	
	\twocolumn[
	\begin{@twocolumnfalse}
		\begin{center}
			\noindent{\Large{\textbf{Sculpt3D: Multi-View Consistent Text-to-3D Generation with Sparse 3D Prior: Supplementary Material}}}
		\end{center}
		\vspace{0.4in}
	\end{@twocolumnfalse}
	]

\section{Supplementary Materials}
We have prepared supplementary materials, including a document and a video, to provide a more comprehensive understanding of our Sculpt3D.
In the document, we discuss the technical details of our implementation in~\cref{sec:tech_details}.
Moreover, we present additional examples and comparisons in~\cref{sec:more_res} to demonstrate the performance of our method.
Furthermore, we have prepared a video that showcases the results and comparisons of Sculpt3D. 


\section{Technical Details}
\label{sec:tech_details}

\paragraph{Implementation details.} This section provides more implementation details of our experiments. 

\begin{itemize}
    \item In our experiments, we observe that most objects in Objectverse \cite{deitke2023objaverse} are aligned with the observer's frontal view. Thus we only normalized the vertices and centers of the templates without manually adjusting their poses. In addition to the 100 prompts provided by T3bench, we also use ChatGPT to generate 40 additional prompts, including common objects as well as some unusual and special items. The prompts we used with ChatGPT are as follows.   
    \textit{I am utilizing a text-to-3D model to generate various 3D objects. Please create 40 prompts for me, including both everyday common objects and some unusual and special items. }
    
    \item 
    When performing appearance modulation, three adapters are utilized to correct erroneous patterns without altering the style of the generated objects. The adapters we used are a spatial color palette adapter, a structure adapter, and an image adapter. Specifically, since T2I-Adapter \cite{mou2023t2i} supports combining multiple adapters to utilize complementary ability between different adapters, we use sketches extracted from the template objects as structure pattern conditions and the hue and color distribution of the generated object as spatial color conditions. Both the sketch and spatial color palette extraction model are the default models from the T2I-Adapter. This approach effectively retains the pattern information of the template while transferring it to the color distribution of the generated object.
    When the color distribution of the template is transferred, it is used as the image condition to modulate the diffusion process using the equation \ref{eq_am}.
\end{itemize}

\section{More Results }

\label{sec:more_res}
\paragraph{Loss balancing ablation}
In addition to the ablation studies of the shape learning provided in Sec. \ref{sec:method_ablation}, we conduct another ablation study to discuss the effectiveness of sparse ray sampling, which is described in Sec. \ref{sec:method_shape_learning}.
We first remove sparse ray sampling and keep the value of $\lambda$ in equation \ref{eq3} as 0.1 to evaluate the effectiveness of sparse ray sampling.

\begin{figure}[ht]
    \centering
    \includegraphics[width=1.0\linewidth]{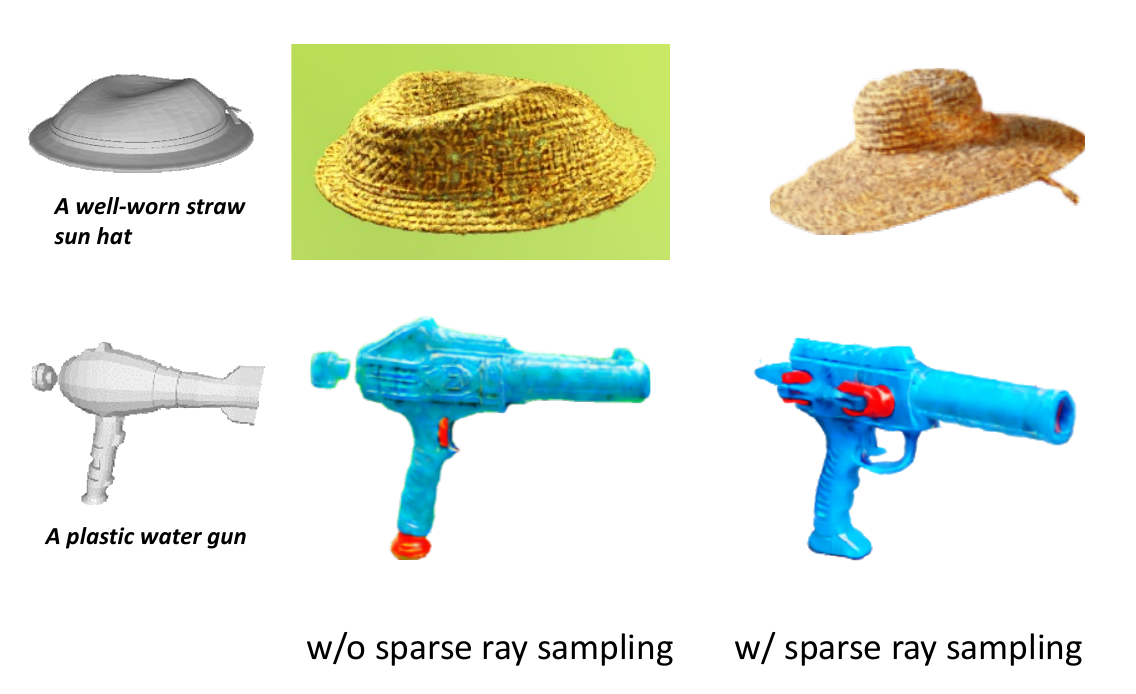}
    \caption{Ablation on the sparse ray sampling strategy.}
    \vspace{-2mm}
    \label{fig_sparse}
\end{figure}

As shown in Figure \ref{fig_sparse}, the results show that removing sparse ray sampling causes the generated objects to closely resemble the template, due to the geometric constraints being uniformly applied to all points. For example, the folds in the hat closely match those in the template, and the back cover of the water gun doesn't close. As shown in the third column of Figure \ref{fig_sparse}, by implementing sparse ray sampling our method can generate imaginative and reasonable geometry under the guidance of the reference shape.

\begin{figure}[ht]
    \centering
    \includegraphics[width=1.0\linewidth]{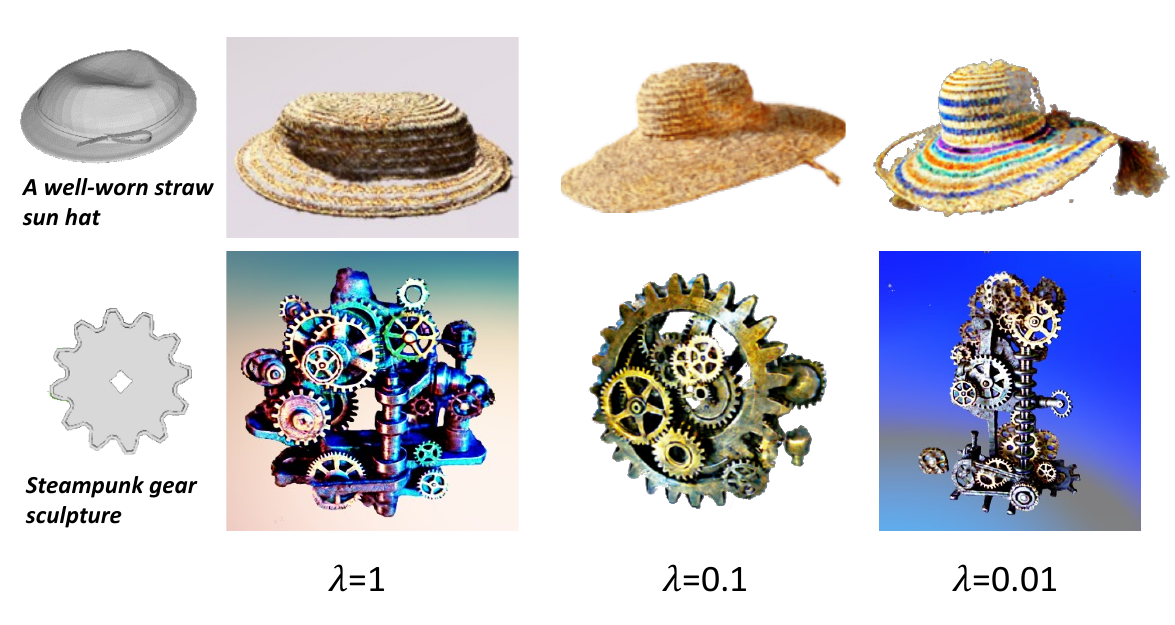}
    \caption{Ablation on the shape co-supervision value $\lambda$ in equation \ref{eq3}.}
    \vspace{-2mm}
    \label{fig_lambda}
\end{figure}

For the choice of  $\lambda$ in equation \ref{eq3}, we study the effect of it by applying sparse ray sampling with $\lambda$ values of 1, 0.1, and 0.01. The results are shown in Figure \ref{fig_lambda}. It's evident that even at $\lambda = 1$, our sparse sampling approach is able to provide sufficient flexibility for the model to learn new shapes. Compared to the results with $\lambda = 1$, setting $\lambda$ as 0.1 can further increase the geometry freedom in the generated results. For instance, the shape of the straw hat is obviously changed. When set $\lambda$ as 0.01, the model can create significantly new geometries, but it may produce undesirable outcomes. Therefore, we default $\lambda$ as 0.1 in our experiments.

\paragraph{Details of comparison with baselines.}

To further validate the effectiveness of the sparse prior scheme in sculpt3d, we conduct two additional experiments.  We first study the effectiveness of the sketch shape loss proposed by Latent-NeRF~\cite{metzer2023latent}. They propose it to allow the model to make slight changes in the template's surface, the loss is formulated as:
\begin{equation}
L_{\text{Sketch-Shape}} = CE(\alpha_{\text{NeRF}}(p), \alpha_{\text{GT}}(p)) \cdot (1 - e^{-\frac{d^2}{2\sigma_S}}),
\end{equation}
where $\alpha_{\text{NeRF}}$ and $\alpha_{\text{GT}}$ are the NeRF occupancy and template shape's occupancy, respectively.
The loss is applied to all points, $d$ represents the distance of a point $p$ from the surface, and $\sigma_S$ is a hyperparameter that controls how lenient the loss is. A higher $\sigma_S$ value means a more relaxed constraint to the surface of the generated object. Since their method operates with the SDS loss at a low resolution of 64 rendering, for a more comprehensive comparison,  we use their code to conduct experiments in their 64 setting and combine it with the VSD loss to train at a higher resolution of 512 rendering. To fully utilize the new geometry generation capability of their method, we employ the maximum value of $\sigma_S$, 1.2, as used by them in all experiments.

\begin{figure}[ht]
    \centering
    \includegraphics[width=1.0\linewidth]{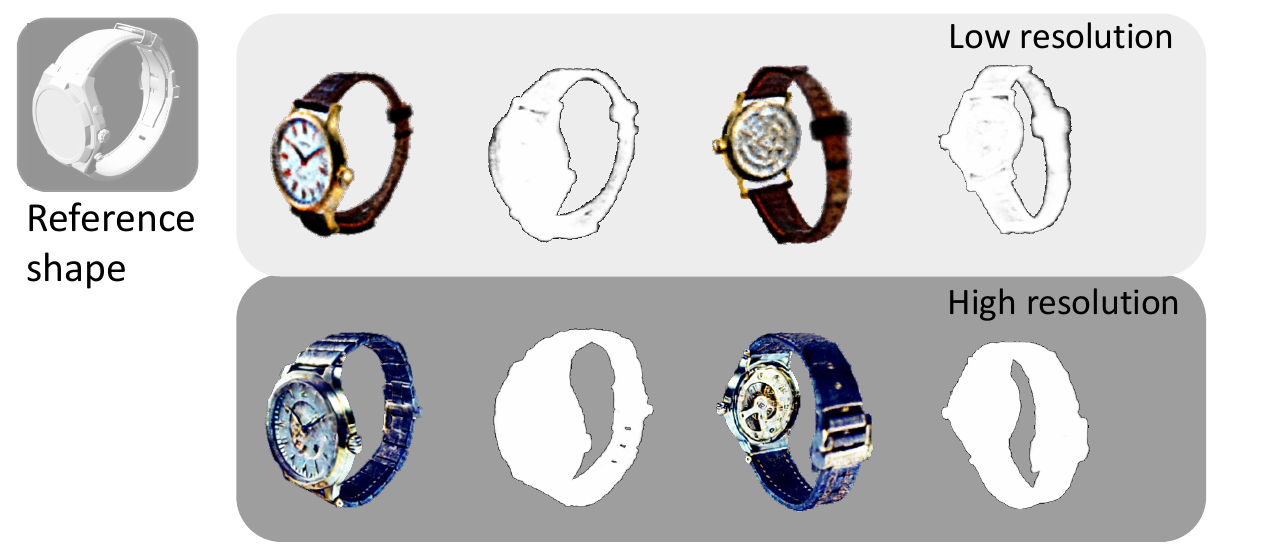}
    \caption{Ablation of the strategy proposed by \cite{metzer2023latent} in both low and high resolution rendering.}
    \vspace{-2mm}
    \label{fig_latentnerf}
\end{figure}

The experimental results are illustrated in Figure \ref{fig_latentnerf}, where we showcase the generated outcomes at two different resolutions along with their corresponding occupancy. It is observable that at lower resolutions, their method is able to alter surfaces, like thinning watch straps. However, at higher resolutions, their approach struggles to change the object's shape, which results in the generated geometries closely resembling the templates.  Additionally,  it is noted that their method of relaxing surface constraints often leaves residual artifacts on the surface. This is evident in the occupancy results of the watch straps in the first row, stemming from an incomplete removal of surface density.

\begin{figure}[ht]
    \centering
    \includegraphics[width=1.0\linewidth]{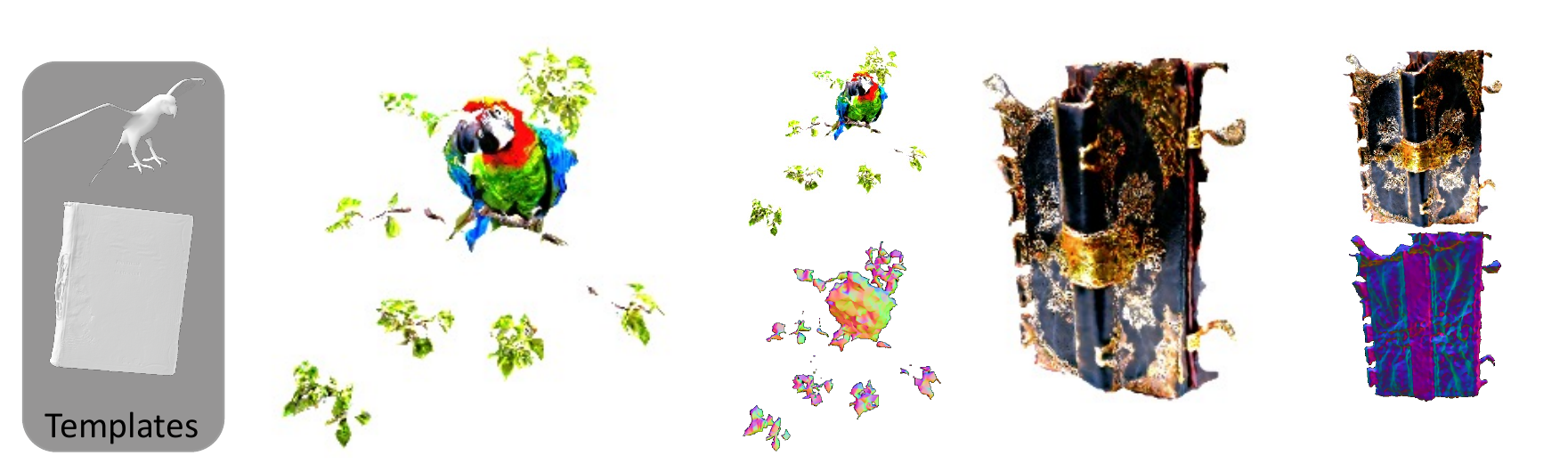}
    \caption{Ablation on the mesh initiation strategy.}
    \vspace{-2mm}
    \label{fig_f3d}
\end{figure}

\paragraph{Comparison with mesh initiation. } In the main text, we mentioned that directly using the template's shape to initialize NeRF's density can not guarantee a satisfactory shape. Unlike our approach, Fantasia3D uses a mesh-based DMTet as a 3D representation, thus supporting initialization with an initial mesh. To more comprehensively verify the role of geometric initialization, we also used our template to initialize Fantasia3D. As shown in Figure \ref{fig_f3d}, the results show that simple initialization is hard to ensure the subsequent learning direction of the model. Despite the model being initialed by a reasonable shape, it still produces unsatisfactory outcomes, such as the distorted shapes of birds and books. This further underscores the necessity of employing geometry and 2D co-supervision during the learning process.

\paragraph{More comparisons.}
\begin{figure*}[h]
    \centering
    \includegraphics[width=0.95\linewidth]{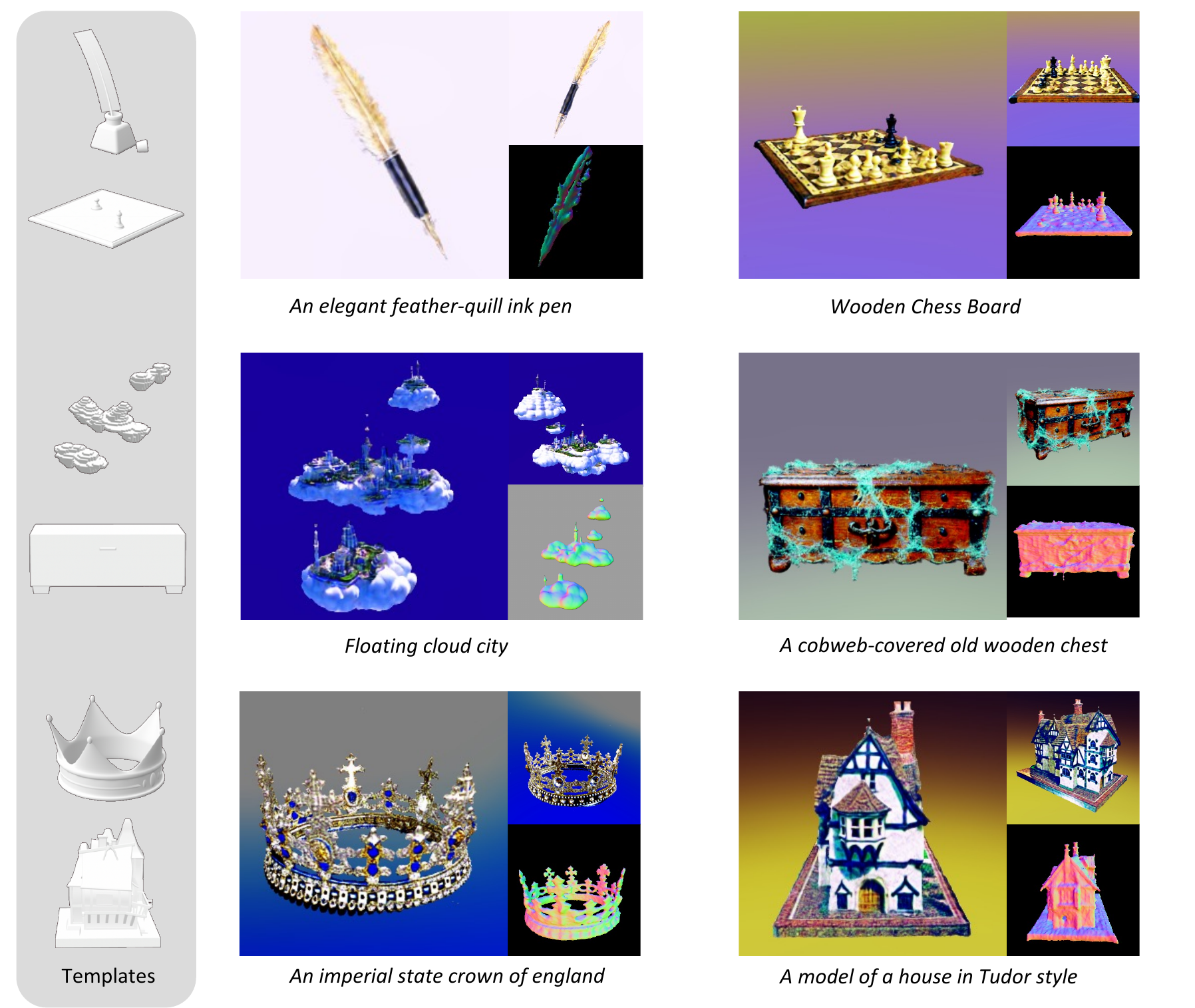}
    \caption{More multi-view examples generated by our method, the retrieved templates are shown on the left.}
    \label{fig:supp_more_results}
\end{figure*}
Here,  we showcase more multi-view examples generated by our method in Figure \ref{fig:supp_more_results}.

\begin{figure*}[p]
    \centering
    \includegraphics[width=0.95\linewidth]{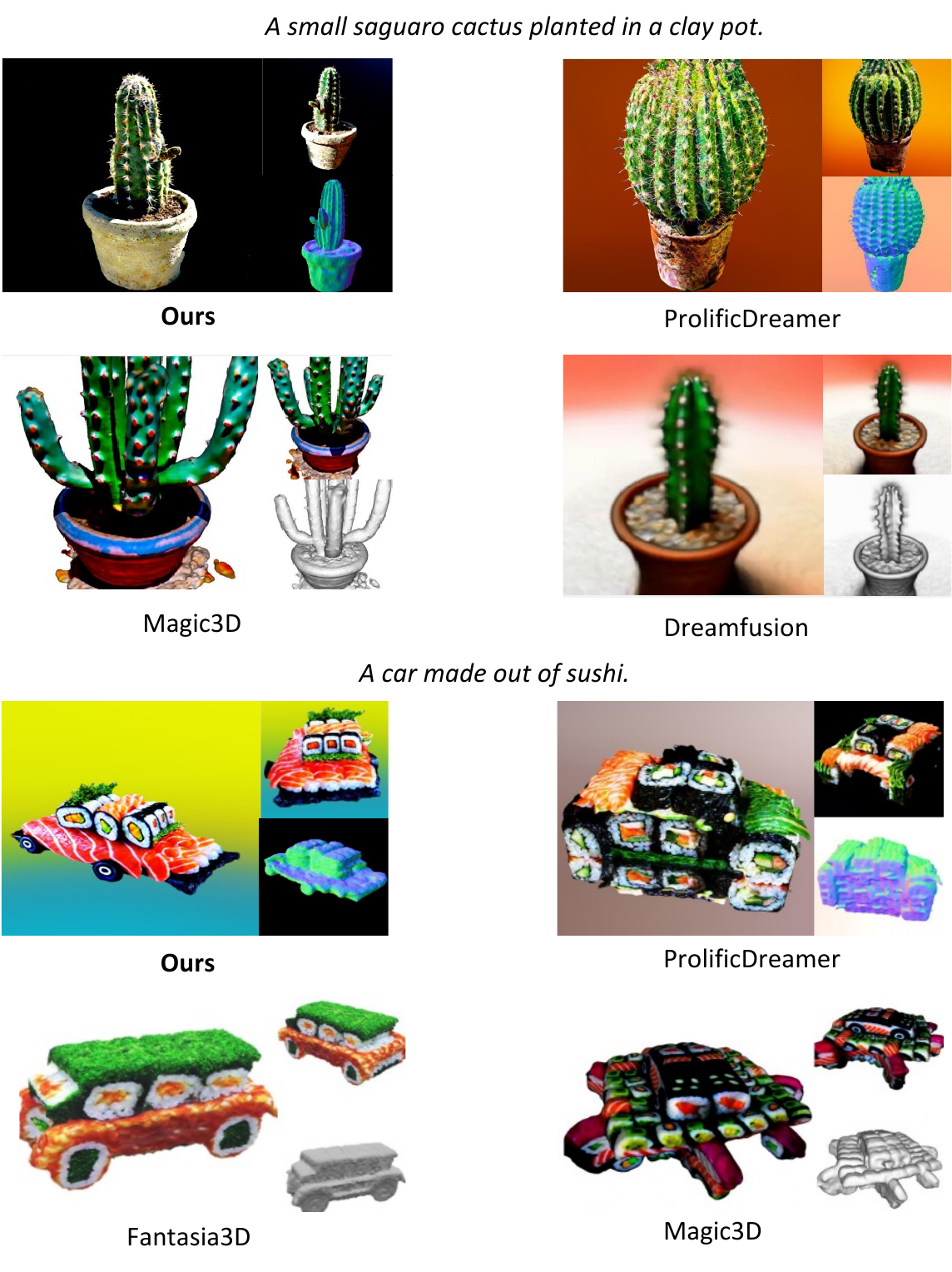}
    \caption{Additional examples for qualitative comparison with baselines.}
    \label{fig:supp_compare}
\end{figure*}

\begin{figure*}[p]
    \centering
    \includegraphics[width=0.95\linewidth]{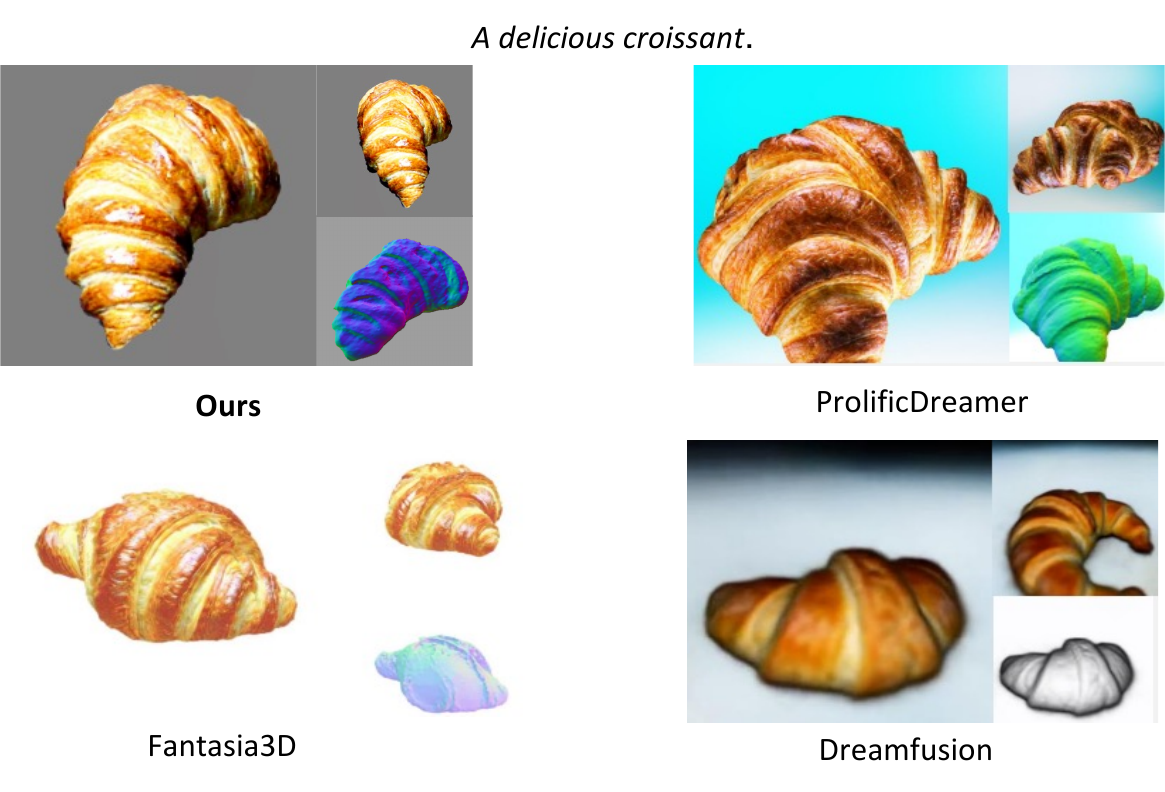}
    \caption{Additional examples for qualitative comparison with baselines.}
    \label{fig:supp_compare_1}
\end{figure*}

Furthermore, we also provide more comparisons with baselines in Figure \ref{fig:supp_compare} and Figure \ref{fig:supp_compare_1}. To compare with the best results demonstrated by the baseline methods, we follow the previous works \cite{wang2023prolificdreamer, chen2023fantasia3d, lin2023magic3d} to directly copy the figures from the corresponding papers for comparisons.

\clearpage


\end{document}